\pdfoutput=1
\documentclass[11pt]{article}
\usepackage[]{acl}
\usepackage{times}
\usepackage{amsfonts}
\usepackage{amsmath}
\usepackage{latexsym}
\usepackage{multirow}
\usepackage{enumerate}
\usepackage{graphicx} 
\usepackage{float}
\usepackage{subfigure}
\usepackage{booktabs}
\usepackage{times}
\usepackage{latexsym}
\usepackage{enumitem}

\usepackage[T1]{fontenc}
\usepackage[utf8]{inputenc}

\usepackage{microtype}

\title{ECO v1: Towards Event-Centric Opinion Mining}

\author{Ruoxi Xu${}^{1,3}$,
Hongyu Lin${}^{1}$\thanks{~ Corresponding Authors},
Meng Liao${}^{4}$\footnotemark[1],
Xianpei Han${}^{1, 2}$, \\
\textbf{
Jin Xu${}^{4}$,
Wei Tan${}^{4}$,
Yingfei Sun${}^{3}$,
Le Sun${}^{1,2}$ }\\
${}^{1}$Chinese Information Processing Laboratory ~ 
${}^{2}$State Key Laboratory of Computer Science \\
Institute of Software, Chinese Academy of Sciences, Beijing, China\\
${}^{3}$School of Electronic, Electrical and Communication Engineering, \\ University of Chinese Academy of Sciences, Beijing, China \\
${}^{4}$Data Quality Team, WeChat, Tencent Inc., China \\
{\tt \{ruoxi2021,hongyu,xianpei,sunle\}@iscas.ac.cn} \\
\tt \{maricoliao,jinxxu,vinnietan\}@tencent.com \\
\tt yfsun@ucas.ac.cn}

\begin{document}
\maketitle
\begin{abstract}
Events are considered as the fundamental building blocks of the world. Mining event-centric opinions can benefit decision making, people communication, and social good. Unfortunately, there is little literature addressing event-centric opinion mining, although which significantly diverges from the well-studied entity-centric opinion mining in connotation, structure, and expression. In this paper, we propose and formulate the task of event-centric opinion mining based on event-argument structure and expression categorizing theory. We also benchmark this task by constructing a pioneer corpus and designing a two-step benchmark framework. Experiment results show that event-centric opinion mining is feasible and challenging, and the proposed task, dataset, and baselines are beneficial for future studies.
\end{abstract}

\section{Introduction}

% 任务的必要性：事件的重要性
Events are the fundamental building blocks of the world \citep{russell1927analysis, ong1969world}.
We express, share and propagate our opinions about events with personal understandings, emotions and attitudes in our daily life. People can better understand, communicate and interact with each other by mining, sharing and exchanging event-centric opinions. And being exposed to event-centric opinions from different angles can debias people's own emotions and  attitudes about social issues \citep{karamibekr2013sentence}. Therefore, mining event-centric opinions have huge social and personal impacts.

\begin{figure}[!ht]
    \centering
    \setlength{\belowcaptionskip}{-20pt}
    \includegraphics[width=0.45\textwidth]{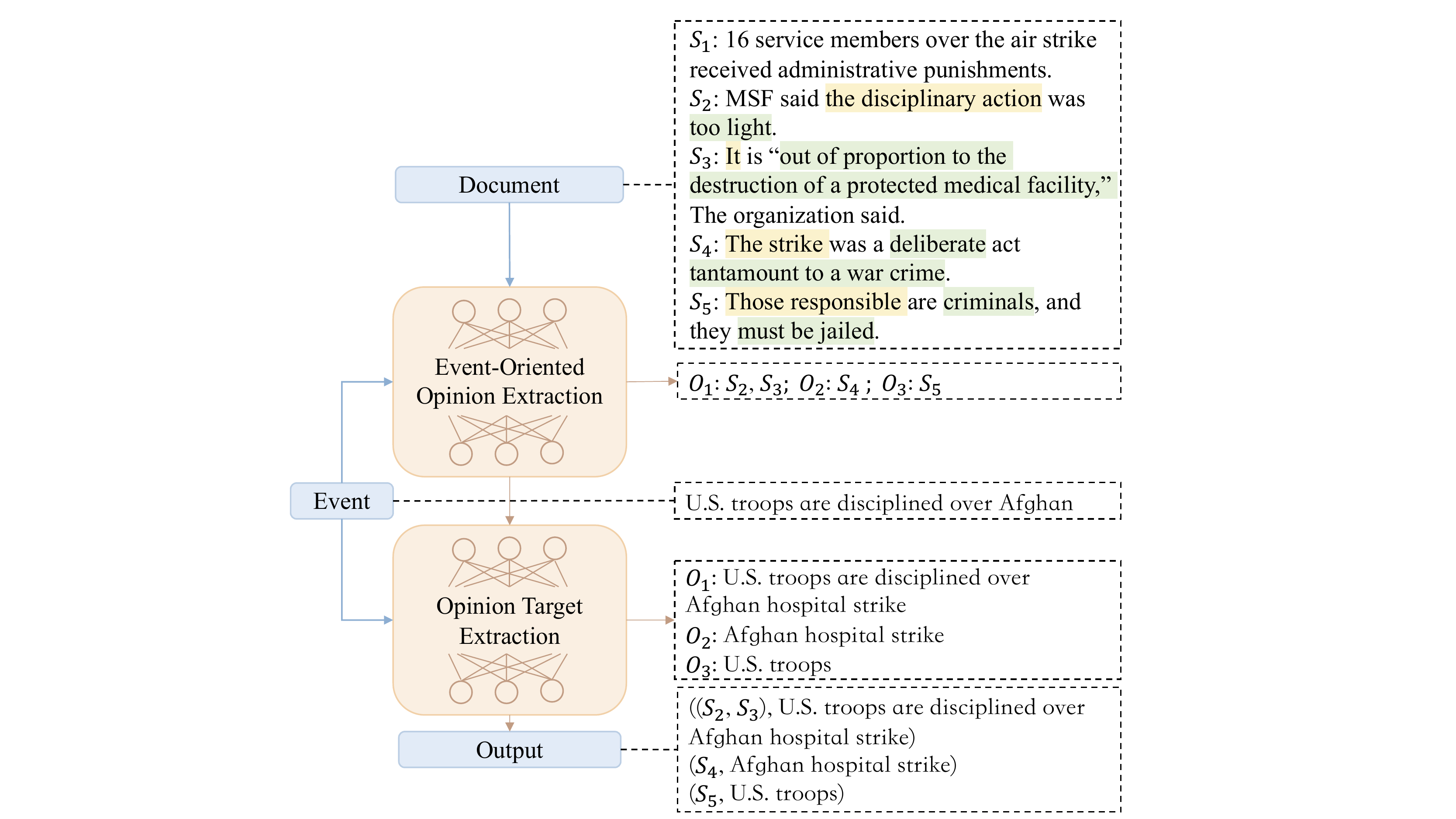} \\
    \caption{An illustration of event-centric opinions. Given an event, people can express their \textcolor[RGB]{169,209,142}{\textbf{judgements}}, \textcolor[RGB]{187,206,228}{\textbf{beliefs}}, \textcolor[RGB]{241,208,168}{\textbf{attitudes}} and \textcolor[RGB]{223,219,229}{\textbf{suggestions}}. The opinions oriented to an event may not directly target at itself, but can target at its related subevents or entities. }
    \label{example}
\end{figure}

Unfortunately, there is little literature addressing \textit{event-centric opinion mining}, and most of current opinion mining studies focus on entity-centric opinions, which significantly diverge from the concerning event-centric ones. 
First, entity-centric opinions mostly focus on sentimental polarity of the holder~\citep{liu2012sentiment}, meanwhile event-centric opinions care more about the content such as non-sentimental judgments, predictions or suggestions. 
Second, due to the rich interactions between events, entities, and people, event-centric opinions have a complicated structure. Given an event, people can express their opinions about the event itself, as well as its subevents, related events, and the involved entities. 
Third, the expressions of event-centric opinions are unique. The targets of event-centric opinions are frequently implicitly referred to, which often don’t appear in the opinion expressions.
Moreover, event-centric opinions are usually widely spread in long news and passages, which are mixed up with facts and non-opinion information. By contrast, entity-centric opinions mainly appear in short and focused reviews or comments individually. 
The above connotation, structure, and expression divergences make event-centric opinion mining a novel task, which cannot be resolved using current entity-centric mining techniques.

\begin{figure}
    \centering
    \setlength{\belowcaptionskip}{-20pt}
    \includegraphics[width=0.48\textwidth]{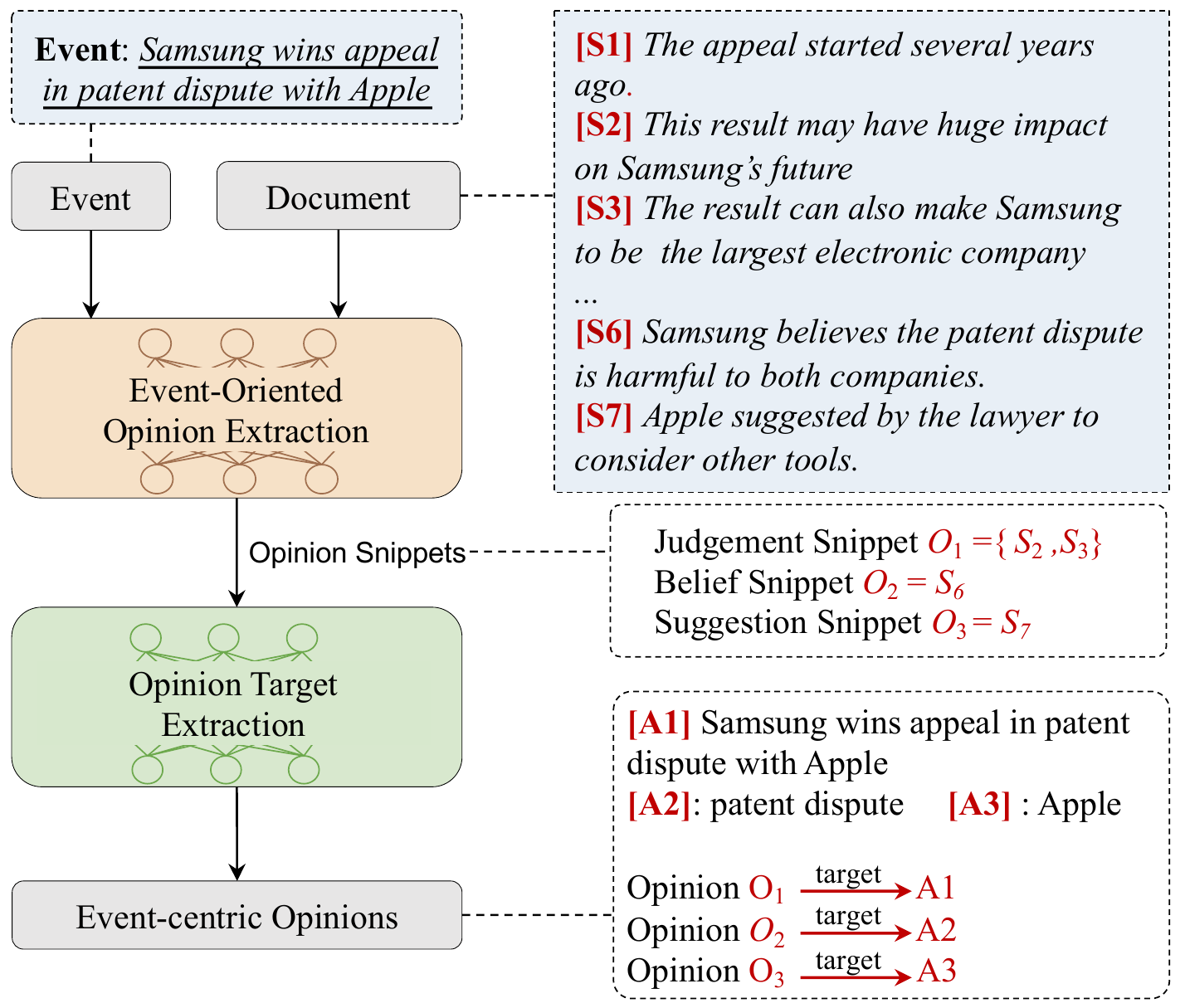} \\
    \caption{Overall architecture of our framework. Here $S$ represents sentence in the document.}
    \label{overall_architecture}
\end{figure}

In this paper, we formally formulate the task of event-centric opinion mining.
Specifically, inspired by the expression categorizing theory~\citep{asher2009appraisal}, we define 5 types of event-centric opinions, and a text snippet is considered as an event-centric opinion if it contains judgments, attitudes, beliefs, sentiments or suggestions of the opinion holder. Then we formulate the targets of event-centric opinions as an \textit{event-arguments opinion structure} by extending the widely-used event-arguments structure~\citep{pustejovsky1991syntax}. In this way, an opinion can target an event itself, or its specific arguments including subevents or involved entities. For example in Figure \ref{example}, given an event ``Samsung wins appeal in patent dispute with Apple'', an opinion towards this event may target at the \texttt{wins appeal} event itself, the related sub-event \texttt{patent dispute}, as well as the involved entity \texttt{Samsung} and \texttt{Apple}. Consequently, event-centric opinion mining can be formulated as identifying opinion snippets from event-related documents and then recognizing the target argument of the opinion snippet.

Based on the task formulation, we create Event-Centric Opinion Bank (ECO Bank), a pioneer corpus for learning and evaluating event-centric opinion mining models. ECO Bank contains nearly 1K events from real-world event trending services, as well as 5K documents about these events in English and Chinese. Each document is aligned to one event. Given a document and its related event, we manually annotated the opinion segments corresponding to the event in the document, and align them to correct target arguments of the event. Consequently, we obtain nearly 18K opinion segments from 5K documents, which target more than 4K different arguments of 1K events.

Finally, we propose a new framework to tackle event-centric opinion mining and benchmark the task on ECO Bank. The overall architecture of the framework is shown in Figure~\ref{overall_architecture}.  Specifically, we decouple event-centric opinion mining into a two-step pipeline. Step 1 is \emph{event-oriented opinion extraction (EOE)}, which detects the snippets containing event-oriented opinions in each document given the concerning event. Step 2 is \emph{opinion target extraction (OTE)}, which recognizes the corresponding target arguments in the event given identified opinion snippets. We then provide two baselines for each step. For event-oriented opinion extraction, we formulate it as either a sentence-level sequential labeling task or a binary sentence classification task. For opinion target extraction, we resolve it based on a span ranking model or an MRC model. By comparing and analyzing the performance of different baselines, we figure out the critical challenges and bottlenecks of current methods to event-centric opinion mining, which can shed some light on the future research directions in this field.

Generally, the contributions\footnote{ECO Bank and the source code are available at \href{http://www.e-com.ac.cn/}{e-com.ac.cn}.} of this paper are:
\begin{itemize}[leftmargin=18pt,topsep=2pt,itemsep=1pt,parsep=1pt]
    \item We propose, define and formulate the task of event-centric opinion mining based on event-argument structure and expression categorizing theory. To the best of our knowledge, this is the first work that tries to formally formulate event-centric opinions and the task of event-centric opinion mining.
    \item We construct Event-Centric Opinion Bank (ECO Bank), a pioneer corpus for learning and benchmarking event-centric opinion mining models in both English and Chinese. To the best of our knowledge, this is the first public benchmark focusing on event-centric opinions.
    \item We design a two-step framework to tackle event-centric opinion mining, and propose several baseline approaches to identify and analyze the challenges and bottlenecks of the task.
\end{itemize}

\section{Event-Centric Opinion Mining}
This section first defines the connotation and targets of event-centric opinions. Then we will formulate the task of event-centric opinion mining.

\subsection{Connotation of Event-Centric Opinions}
The connotations of event-centric opinions are complicated and cannot be simply summarized based on sentimental tendencies. For example, towards a \textit{Trade War} event, one may express the personal judgment opinion by commenting ``\emph{I do believe this is a turning point of the relationships between two countries}'', which is without explicit sentiments. Therefore, we need to define broader connotations for event-centric opinions than entity-centric ones.

To this end, we formulate the connotations of event-centric opinions according to the divergence between facts and opinions~\citep{banfield1984unspeakable, hackett1984decline}.
An event-centric opinion is defined as a statement that expresses views about an event or related issues, which 1) cannot be proved or disproved with currently available information and 2) varies from person to person~\citep{schauer1978language, wiebe2005annotating, corvino2014fact}.

Specifically, inspired by expression categorizing theory ~\citep{asher2009appraisal}, we define the connotation of event-centric opinion as a text snippet that expresses the following 5 kinds of information, including: 1) \textbf{Judgements}, such as speculations, interpretations, and predictions about things in the future, e.g., \textit{Trump's plan will not work}; 2) \textbf{Attitudes}, such as positions on controversial issues and evaluations of people, places, and things, e.g., \textit{Apple describes the ruling as total political crap}; 3) \textbf{Sentiments}, which express feelings like fear and sadness, e.g., \textit{I am so happy to see the Act passed}; 4) \textbf{Beliefs}, which can not be proved or disproved, e.g. \textit{I believe aliens definitely exist}; 5) \textbf{Suggestions}, which is about personal advice to the readers, e.g., \textit{We advise Samsung Galaxy Note 7 owners to turn off their devices during flights}.

\subsection{Targets of Event-Centric Opinions}
Entity-centric opinions mostly directly target the entity or its attributes (which is referred as \textit{aspects}). By contrast, when talking about events, people can talk about their related events, entities and concepts, and these opinions may not direct to the event itself. 
For example, given the \textit{Trade War} event, an opinion holder may express their opinion like \emph{Trump always made bad decisions}, which actually targets on \emph{Trump} rather than the event because the holder will not change the opinion no matter whether the \textit{Trade War} event happens. 

To formulate the target of event-centric opinions, we introduce event-arguments opinion structure. Event-arguments structure~\citep{pustejovsky1991syntax} is a widely used event formulation in many event-related tasks where arguments refer to a set of critical elements about how an event realized~\citep{doddington2004automatic,hovy2013events}.
Based on this structure, the opinions about a specific event target at one of the following arguments: 1) \textbf{Event}, which means that the opinion is directly targeting the entire event; 2) \textbf{Subevents}, which means that the opinion does not target at the entire event, but on its subevent or related event. For example in Figure~\ref{example}, an opinion towards the \emph{patent dispute} in the event \emph{Samsung wins appeal in patent dispute with Apple}; 3) \textbf{Entities}, which means that the opinion directly targets one of the involved entities regardless of the event, e.g., commenting \textit{Apple is a great company} on the event in Figure~\ref{example}.

\subsection{Task Formulation for Event-Centric Opinion Mining}
Based on the formulated connotation and targets, we define event-centric opinion mining as the task of extracting (opinion, argument) pairs from a document and an event descriptor. Formally, let $e=\{w_1,w_2,...,w_m\}$ denote an event descriptor with $m$ tokens and $d = \{s_1,s_2,...,s_n\}$ denote a document with $n$ sentences.
Event-centric opinion mining aims to identify (opinion, argument) pairs $T=\{...,(o_k,a_k),...|e,d\}$, where $o = \{s_i,s_{i+1},...,s_j| s \in d\}$ is a continuous opinion segment in $d$ targeting at the same argument, and $a = \{w_t,w_{t+1},...,w_l| w\in e\}$ is the target argument of the opinion $o$ in the event descriptor $e$. For example in Figure~\ref{example}, given the event \emph{Samsung wins appeal in patent dispute with Apple}, there are 4 (opinion,argument) pairs, two of whom target at the entire event, one target at subevent \emph{patent dispute} and one target at entity \emph{Apple}.

\section{Event-Centric Opinion Bank}
Based on the task formulation, this paper creates Event-Centric Opinion Bank (ECO Bank), a new event-centric opinion mining corpus in both English and Chinese. In the following, we describe how we construct ECO Bank and report the statistics of ECO Bank.

\subsection{Dataset Construction}
\noindent\textbf{Event Descriptor Collection.} To construct ECO Bank, we collect event descriptors from real-world event trending services. For the English portion of ECO Bank, we collect event descriptors from the W2E dataset~\citep{hoang2018w2e}, a worldwide event dataset for topic detection and tracking. We select highly discussed topics as event descriptors and manually shorten them into meaningful texts if necessary. For the Chinese portion, we collect manually-maintained event trending from widely-used social networks, WeChat Top Topics. We then filter out items in the trending corresponding to events as event descriptors. Because the trending is manually created, it is already of sufficient quality and therefore no more modification is required. Finally, we construct 988 high-quality event descriptors, where 821 in the Chinese and 167 in the English.

\noindent\textbf{Document Collection.} Given the event descriptor, we collect related documents that may contain the opinions towards the event. For the Chinese portion of ECO Bank, we collect related documents by retrieving relevant documents from WeChat Search. Specifically, We retrieve top 10 articles for each event descriptor, and then manually filter out the redundant, low-quality and irrelevant documents. For the English portion, because each topic in W2E dataset is already linked to several related articles from more than 50 prominent mass media channels. We therefore directly applied these documents except filtering out the redundant ones. Finally, we preserve 3000  Chinese documents and 2000 English documents for further annotation.

\noindent\textbf{Event-Centric Opinion Annotation.} Given the documents and their corresponding event descriptors, we hired annotators to annotate the (opinion, argument) pairs. Specifically, the annotation is conducted in a two-step paradigm. First, annotators are asked to identify opinions related to the event described in the descriptor. Then, given the event descriptor, an identified opinion and the source document, the annotators were asked to link the opinion to its target in the descriptor.
To ensure the high quality of annotations, each document is annotated by two annotators. If there is a disagreement between the original annotators, three more professional annotators will relabel the document independently, and produce the final annotations by voting between them. Finally, to facilitate further research, we also ask annotators to recognize all possible arguments in the event descriptor that can serve as an opinion target. All annotators are fairly paid according to their workload.

\subsection{Dataset Analysis}
Table~\ref{dataset} shows the main statistics of ECO Bank. We can observe many unique characteristics of event-centric opinion mining. First, the distribution of event-centric opinions is very sparse. Only about 30\% of the sentences in both English and Chinese dataset express opinions. This is because event-centric articles usually mix massive factual snippets with opinionated snippets. By contrast, entity-centric opinions are densely distributed in comments and reviews. Second, the targets of event-centric opinions are highly diversified. We notice that only ~30\% of opinions directly target the event, leaving 24.9\% on subevents and 44.5\% on entities (ECO-ZH), and 11.7\% on subevents and 53.9\% on entities (ECO-EN). This verifies the necessity of defining an event-specified opinion structure. Furthermore, we find that targets of event-centric opinions are often implicit. To show this we randomly select 50 documents with 151 opinions. Among them, there are 80 opinions on events/subevents, where 25\% opinions target implicit arguments, and 28\% opinions are with event co-reference and therefore its target event cannot be directly recognized without more contexts. By contrast, this proportion is much lower in opinions on entity arguments, where we only find 8\% implicit arguments and 7\% entity co-reference. These results demonstrate that the target of event-centric opinions cannot be identified locally, which is one of the most significant divergences between event-centric and entity-centric opinion mining.

\begin{table}[!t]
\centering
\setlength{\belowcaptionskip}{-10pt}
\resizebox{0.49\textwidth}{!}{
\begin{tabular}{|c|c|c|c|}
\hline
\multicolumn{2}{|c|}{\textbf{Statistics on ECO Bank}} & \textbf{Chinese} & \textbf{English}\\
\hline
\multirow{3}*{Document} & Number & 3000 & 2000\\
\cline{2-4}
& Avg. Sents & 15.2 & 20.3 \\
\cline{2-4}
& Avg. Opinion & 2.6 & 4.5 \\
\hline
\multirow{3}*{Opinion} & Number & 7742 & 9058\\
\cline{2-4}
& Ratio (\%) & 32.0 & 28.1\\
\cline{2-4}
& Avg. Sents & 1.9 & 1.3\\
\hline
\multirow{2}*{Event} 
& Number & 821 & 167\\
\cline{2-4}
& Avg. Tokens & 6.4 & 7.7\\
\hline
\multirow{3}*{Arguments} & Events (\%) & 30.6 & 34.4\\
\cline{2-4}
& Subevents (\%) & 24.9 & 11.7\\
\cline{2-4}
& Entities (\%) & 44.5 & 53.9\\
\hline
\end{tabular}
}
\caption{Overall statistics of ECO Bank dataset.}
\label{dataset}
\end{table}

\section{Benchmarking Event-Centric Opinion Mining}
This section benchmarks event-centric opinion mining with a two-step framework. 
Two feasible solutions are proposed for each step, and therefore lead to 4 different benchmark architectures.

\subsection{Step 1: Event-Oriented Opinion Extraction}
Given an event descriptor $e$ and a related document $d$, the goal of event-oriented opinion extraction (EOE) is to extract text snippets $o$ in $d$ which contain opinions about event $e$. To this end, we propose two architectures, one formulates EOE as a pair-wise classification task and the other formulates it as a sentence-level sequential labeling task.

\subsubsection{Pair-wise Classification}
\label{paircls_section}
A basic solution for EOE is to build binary classifier for all (sentence, event) pairs. Specifically, given an event $e$ and a sentence $s$ in document $d$, we identify whether $s$ is an opinion to $e$ using a BERT-based binary classifier~\citep{devlin2018bert}. The classifier takes the concatenation $\mathcal{X} = \{{\rm [CLS]},e,{\rm [SEP]},s\}$ as input,
where [CLS] and [SEP] represent the beginning of input and the separator between $s$ and $e$ respectively. We then use BERT as the encoder, then conduct binary classification on [CLS] token to identify the relation between $e$ and $s$:
\begin{equation}
\begin{aligned}
    \mathbf{\mathcal{H}}= {\rm BERT}(\mathcal{X}), \indent
    p= {\rm sigmoid}(\mathbf{\mathcal{H}}_{{\rm [CLS]}}), \label{eq:sigmod}
\end{aligned}
\end{equation}
where $\mathbf{\mathcal{H}}_{{\rm [CLS]}}$ is the representation at [CLS] token, and $p$ is the probability of $s$ containing an opinion to $e$. Then we regard sentences with $p \leq 0.5$ as the opinion sentences, and concatenates all continuous opinion sentences to form opinion snippets.

\subsubsection{Sentence-level Sequential Labeling}
\label{seq_section}
Because an opinion may contain more than one sentence, sentence-level classification to identify opinion snippets may result in opinion boundary ambiguity. To this end, we propose sentence-level sequential labeling architecture~\citep{cheng2020argument} for EOE. 
Specifically, given a document $d= \{s_1,s_2,...,s_n\}$ and an event descriptor $e$, we first concatenate $e$ and each sentence in $d$ to form the input $\mathcal{X}$, using [CLS] as the separators between each sentence. We then fed $\mathcal{X}$ into BERT-based encoder to obtain context-aware representations. The representations at [CLS] tokens $\mathbf{\mathcal{H}}_{{\rm [CLS]}}$ are used to represent the sentences after them. To leverage the deep interaction between different sentences, we further apply BiLSTM layer upon $\mathbf{\mathcal{H}}_{{\rm [CLS]}}$ and learn the interacted sentence representations $\mathbf{\mathcal{S}} = {\rm BiLSTM}(\mathbf{\mathcal{H}}_{{\rm [CLS]}})$.
Finally, we apply a Conditional Random Field~\citep{lafferty2001conditional} upon $\mathbf{\mathcal{S}}$ to label each sentence to obtain the sentence-level tagging output $\mathbf{\mathcal{Y}} = {\rm CRF}(\mathbf{\mathcal{S}})$ encoded in BIO schema~\citep{sang2000introduction}.

\subsection{Step 2: Opinion Target Extraction}
Given an event descriptor $e = \{w_1,...,w_n\}$ and an opinion snippet $o$ identified in Step 1, Opinion Target Extraction (OTE) aims to recognize a span in $e$ corresponding to the target argument of $o$. To this end, we build two baselines of OTE by taking it as either a span ranking problem or a MRC problem.

\subsubsection{Span Ranking for Opinion Target Extraction}
\label{spanr_section}
Given an opinion $o$, the span ranking approach directly enumerates all spans in $e$, and selects the best span as $o$'s target argument. 
Formally, given a span $a$ in $e$, we concatenate $a$ with $o$ to form the model input. Then similar to the pair-wise EOE classifier in Equation (\ref{eq:sigmod}), we send the concatenation into a BERT-based encoder, and then obtain the score of the span $a$ being the opinion target of $o$ via a sigmoid classifier. Finally, the span with highest score is regarded as the target of the opinion $o$.

\begin{table*}[!t]
\centering
\setlength{\belowcaptionskip}{-10pt}
\resizebox{0.9\textwidth}{!}{
\begin{tabular}{c|ccc|ccc|ccc|ccc}
\toprule
& \multicolumn{6}{c|}{ECO-ZH} & \multicolumn{6}{c}{ECO-EN} \\
& \multicolumn{3}{c}{Segment level} & \multicolumn{3}{c|}{Sentence level}
& \multicolumn{3}{c}{Segment level} & \multicolumn{3}{c}{Sentence level} \\
\midrule
& P & R & $F_1$ & P & R & $F_1$
& P & R & $F_1$ & P & R & $F_1$\\
PairCls-SpanR
& 14.51	& 12.08	& 13.18
&37.69	&33.61	&35.50
&6.86 	&4.83 	&5.76 	
&13.77 	&12.64 	&13.18 
\\
PairCls-MRC 
&13.45	&11.19	&12.22
&48.99	&43.67	&46.13
&14.67	&10.42	&12.19
&33.12	&29.32	&31.10
\\ 
Seq-SpanR
&25.07	&21.77	&23.31
&35.46	&28.07	&31.34
&9.24 	&9.96 	&9.59 	
&11.30 	&12.54 	&11.89 
\\
Seq-MRC
&29.72	&26.48	&28.01
&47.74	&37.80	&42.19
&17.02	&19.44	&18.15
&24.71	&27.77	&26.15
\\
\midrule \midrule
Human 
& 86.96	& 86.02	& 86.49	& 79.46	& 94.23	& 86.22
& 72.59	& 82.10	& 80.83	& 86.78	& 86.07	& 86.42\\
\bottomrule
\end{tabular}
}
\caption{Overall experiment results on ECO-ZH and ECO-EN datasets. \emph{PairCls} denotes pair-wise classification method (\$ \ref{paircls_section}), \emph{Seq} denote sentence-level sequential labeling method (\$ \ref{seq_section}) for EOE, and \emph{SpanR} denotes Span ranker (\$ \ref{spanr_section}), \emph{MRC} denotes MRC method (\$ \ref{mrc_section}) for OTE. We also represent human performance as \emph{Human}.}
\label{overall_performace}
\end{table*}

\subsubsection{MRC for Opinion Target Extraction}
\label{mrc_section}
Recent advances~\citep{cui2020revisiting, sugawara2020assessing} have shown that pointer network style machine reading comprehension models~\citep{wang2016machine} can effectively resolve the span spotting problems. Therefore, we apply an MRC architecture similar to~\citet{devlin2018bert} for OTE, which regards the opinion $o$ as the query and the event descriptor $e=\{w_1,w_2,...,w_n\}$ as the document to identify argument $a$ from $e$.

Specifically, given an event descriptor $e$ and opinion $o$,  we first represent the input $o$ and $e$ as a single packed sequence $\mathcal{X} = \{{\rm [CLS]},o,{\rm [SEP]},e\}$.
We then introduce a start vector $\mathcal{S}$ and an end vector $\mathcal{E}$. The probability of word $w_i$ being the start or end of the argument span is computed as a dot product between $\mathbf{\mathcal{H}_i}$ and $\mathcal{S}$ or $\mathcal{E}$ followed by a softmax over all of words in $e$:
\begin{equation}
\begin{aligned}
    p_s= \frac{e^{\mathcal{S}\mathbf{\mathcal{H}}_i}}{\sum_j e^{\mathcal{S}\mathbf{\mathcal{H}}_j}}, \indent
    p_e= \frac{e^{\mathcal{E}\mathbf{\mathcal{H}}_i}}{\sum_j e^{\mathcal{E}\mathbf{\mathcal{H}}_j}}. \label{eq:mrc_s_sigmoid}
\end{aligned}
\end{equation}
The score of a candidate span from position $i$ to $j$ is defined as $\mathcal{S}\mathbf{\mathcal{H}}_i + \mathcal{E}\mathbf{\mathcal{H}}_j$, and the maximum scoring span where $i \leq j$ is used as a prediction.

\section{Experiments}
\subsection{Benchmark Settings}
\noindent\textbf{Dataset Split.}
We split both English and Chinese portion of Event-Centric Opinion Bank into roughly 7:1:2 for train/dev/test respectively. To ensure no information leakage, the same event descriptor will not be sampled into different sets. Finally, for English portion, there are 112/16/39 event descriptors with 1402/198/400 documents for train/dev/test. And for Chinese portion, there are 590/78/153 event descriptors with 2100/299/601 documents for train/dev/test. This ECO Bank split can be viewed as a standard benchmark for evaluating event-centric opinion mining models.

\noindent\textbf{Evaluation Criteria.}
To evaluate the event-centric opinion mining performance, we design several evaluation metrics for the task as well as its two sub-tasks. Specifically, given golden (opinion,argument) pair set $\mathcal{T}=\{(o^T_1,a^T_1),...,(o^T_n,a^T_n)\}$ and the predicted (opinion,argument) pair set $\mathcal{P}=\{(o^P_1,a^P_1),...,(o^P_n,a^P_n)\}$, where $o_i=\{s_{i1},...,s_{ik}\}$ contains continuous sentences from documents and $a_i=\{w_{i1},..,w_{il}\}$ contains continuous words from the event descriptors, we design the following evaluation metrics:

\textbf{1. End2End Evaluation}, which measures the end-to-end performance of event-centric opinion mining. We propose to use ${\rm F}_1$ score at opinion segment-level or sentence-level to evaluate the overall performance. Segment-level ${\rm F}_1$ is the ${\rm F}_1$ score calculated by directly comparing $\mathcal{T}$ and  $\mathcal{P}$.

And sentence-level ${\rm F}_1$ is calculated by first
splitting (o,a) pairs in $\mathcal{T}$ and $\mathcal{P}$ into sentence-level pairs $\{(s_1,a),..,(s_k,a)\}$ and then combining them to form the sentence-level golden annotation set $\mathcal{T'}$ and prediction set $\mathcal{P'}$. Finally, sentence-level ${\rm F}_1$ is calculated between $\mathcal{T'}$ and $\mathcal{P'}$. % according to Equation (\ref{eq:f1}).

\textbf{2. EOE Evaluation}. We also consider both segment-level and sentence-level metrics when evaluating the Step 1 EOE. The only difference is that we only evaluate the performance of extracting opinion snippets without considering corresponding opinion targets in EOE evaluation.

\textbf{3. OTE Evaluation}. To evaluate how well the Step 2 OTE works, we further use the golden annotated opinion snippets as input to evaluate OTE performance. We use two evaluation metrics for OTE: 1) Accuracy, which measures whether the extracted argument can be exactly the same as the annotated one; 2) Overlap-${\rm F}_1$, which measures the overlap between extracted and golden arguments using ${\rm F}_1$. Specifically, let $a^T=\{w^T_{1},..,w^T_{l}\}$ denotes the golden argument and $a^P=\{w^P_{1},..,w^P_{k}\}$ as the predicted argument, Overlap-${\rm F}_1$ is calculated by micro-averaged $F_1$ on all $(a^T,a^P)$ pairs.

\begin{table}[!t]
\centering
\setlength{\belowcaptionskip}{-10pt}
\resizebox{0.45\textwidth}{!}{
\begin{tabular}{c|cc|cc}
\toprule
& \multicolumn{2}{c|}{ECO-ZH} & \multicolumn{2}{c}{ECO-EN} \\
& Segment-F1 & Sent-F1
& Segment-F1 & Sent-F1 \\
\midrule
PairCls 
&24.39	&67.35
&25.07	&53.40
\\
Seq
&44.33	&62.53
&34.84	&48.41
\\
\bottomrule
\end{tabular}
}
\caption{The performance on Event-oriented Opinion Extraction.}
\label{opinion_performace}
\end{table}

\begin{table}[!t]
\centering
\setlength{\belowcaptionskip}{-10pt}
\resizebox{0.45\textwidth}{!}{
\begin{tabular}{c|cc|cc}
\toprule
& \multicolumn{2}{c|}{ECO-ZH} & \multicolumn{2}{c}{ECO-EN} \\
& Accuracy & Overlap-F1
& Accuracy & Overlap-F1 \\
\midrule
SpanR
&49.21 &77.83
&26.50	&53.31
\\
MRC
& 64.89 & 84.89
&54.29	&76.98
\\
\bottomrule
\end{tabular}
}
\caption{The performance on OTE given golden opinion snippets.}
\label{aspect_performace}
\end{table}

\subsection{Overall Results}
The performance of 4 different architectures on the end2end, EOE and OTE evaluation are shown in Table ~\ref{overall_performace},~\ref{opinion_performace} and ~\ref{aspect_performace}. We also listed the human end2end performance in Table~\ref{overall_performace}, which is summarized from the divergences between the annotations from the first two annotators and the final annotations.
From these tables, we can see that:

\textbf{1) The proposed formulation for event-centric opinion mining is a feasible task for human beings.} From Table~\ref{overall_performace}, we can see that human can reach high agreements on both ECO-ZH and ECO-EN. This demonstrates that the proposed opinion connotation and structure are applicable for event-centric opinions.

\textbf{2) Event-centric opinion mining is a challenging task.} The best benchmark system Seq-MRC can only achieve 28.01 and 18.15 segment-level $F_1$ on Chinese and English respectively. The performance gap between machine and human is huge, which indicates that more effective architectures and task-specialized approaches are needed.

\textbf{3) Seq-MRC architecture achieved the best performance among 4 baseline architectures.} We believe this is because the architecture is a more natural design for event-centric opinion mining. Naturally, EOE is a sentence-level sequential labeling problem given the event, and OTE is a span extraction problem given the opinion. As a result, Seq-MRC is more suitable for solving these two tasks compared with PairCls and SpanR. 

\textbf{4) The main bottleneck for event-centric opinion mining is to identify completed continuous opinion snippets from documents.} From Table~\ref{opinion_performace}, we can see that current sentence-level sequential labeling can only achieve 44.33\% and 34.84\% segment-level $F_1$ score, which is the main reason for the low end2end performance. By contrast, we can see that the sentence-level evaluation results are much better than segment-level evaluation results. We believe the reason behind is that current architectures can not well identify the structural relations at sentence-level, and leveraging such structure requires strong discourse-level knowledge.

\textbf{5) ECO-EN dataset is more challenging than ECO-ZH.} Even with similar training document size and opinion numbers, the performance of ECO-EN is significantly worse than that of ECO-ZH. We believe this is because 1) English opinions are often more implicit than Chinese ones. Therefore, even human annotators made more disagreements on ECO-EN; 2) ECO-EN is with much fewer event descriptors than ECO-ZH, which make the training of EOE models may overfit on the events in the training data.

\subsection{Effects of Opinion Length to EOE}

\begin{figure}
    \centering
    \setlength{\belowcaptionskip}{-10pt}
    \includegraphics[width=0.5\textwidth]{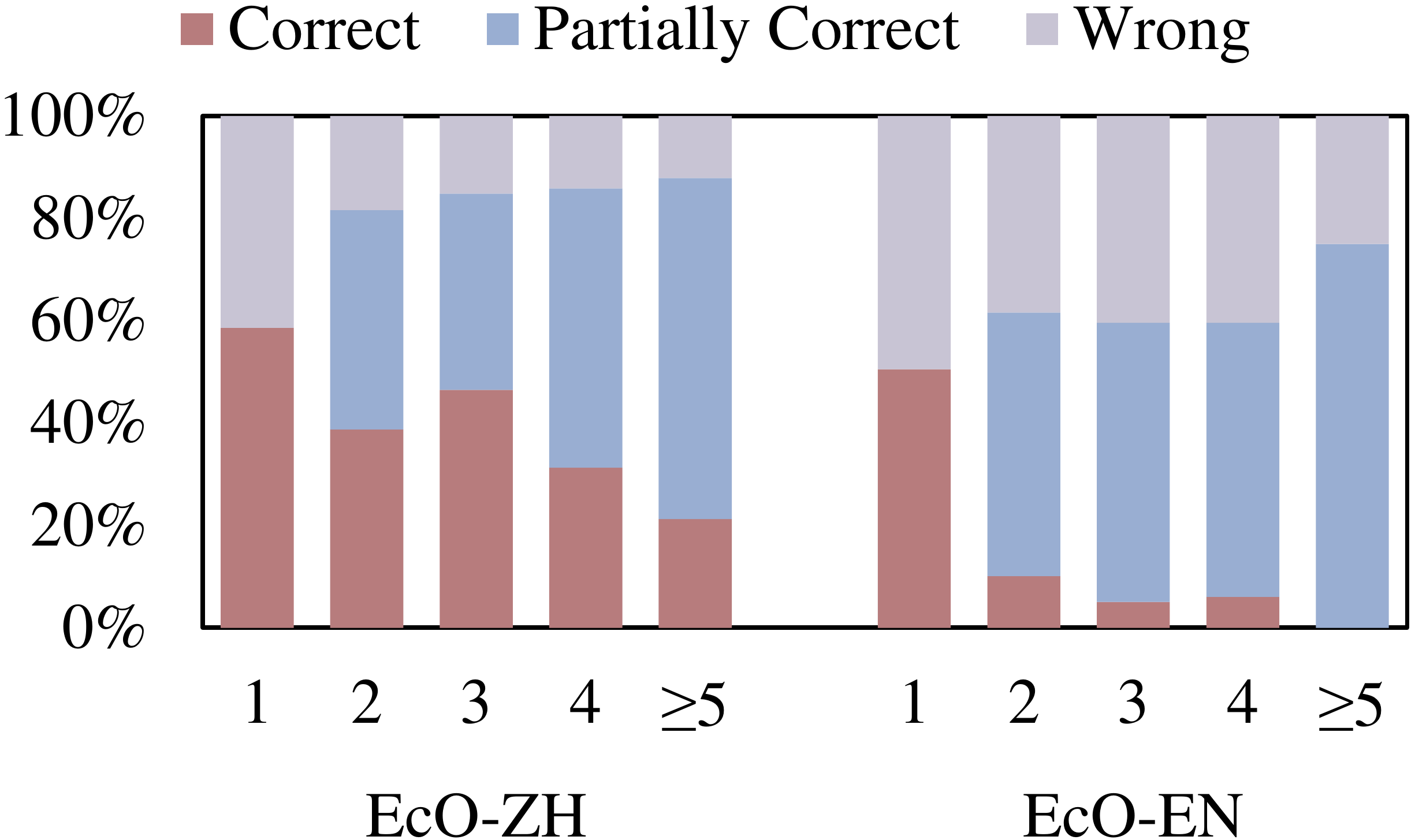} 
    \caption{Performance of \emph{Seq} on EOE with different opinion lengths.}
    \label{opinion_len_analysis}
\end{figure}

To investigate whether opinion length will impact the performance of EOE, we categorize the model's prediction on golden opinion snippets into: 1) Correct; 2) Partially Correct, which means that at least one sentence in the opinion segment is identified; 3) Wrong. Figure~\ref{opinion_len_analysis} shows the results of \emph{Seq} approach. We can see that the correct prediction ratio drops when opinion length increases. This is easy to understand because opinions with more sentences are more difficult to recognize. However, we can see that the wrong prediction ratio also drops along with the increase of opinion length. This indicates that for longer opinions, the chance of at least one sentence can be correctly identified is relatively high. Therefore, if we can jointly consider the predictions of multiple sentences by leveraging discourse knowledge, we may reduce such partial labeling errors and improve the performance.

\subsection{Effects of Argument Type to OTE}

\begin{table}[!ht]
\centering
\resizebox{0.4\textwidth}{!}{
\begin{tabular}{c|ccc}
\toprule
& Subevents & Entities
& Events \\
\midrule
ECO-ZH 
&76.80	&54.40
&72.92\\
ECO-EN
&25.00	&48.48
&70.85\\
\bottomrule
\end{tabular}
}
\caption{Performance of \emph{MRC} on different kinds of arguments.}
\label{comparison_subevents_participants}
\end{table}

Table \ref{comparison_subevents_participants} shows the opinion target extraction performance of \emph{MRC} on different types of arguments. 
For both ECO-ZH and ECO-EN dataset, the performance on whole events is better than that on arguments. In particular, it performs poorly on subevents in ECO-EN. This may be because the amount of event descriptors is not enough for the model to learn to extract the exact boundaries of subevent arguments.

\begin{figure}[!t]
    \centering
    \setlength{\belowcaptionskip}{-15pt}
    \includegraphics[width=0.42\textwidth]{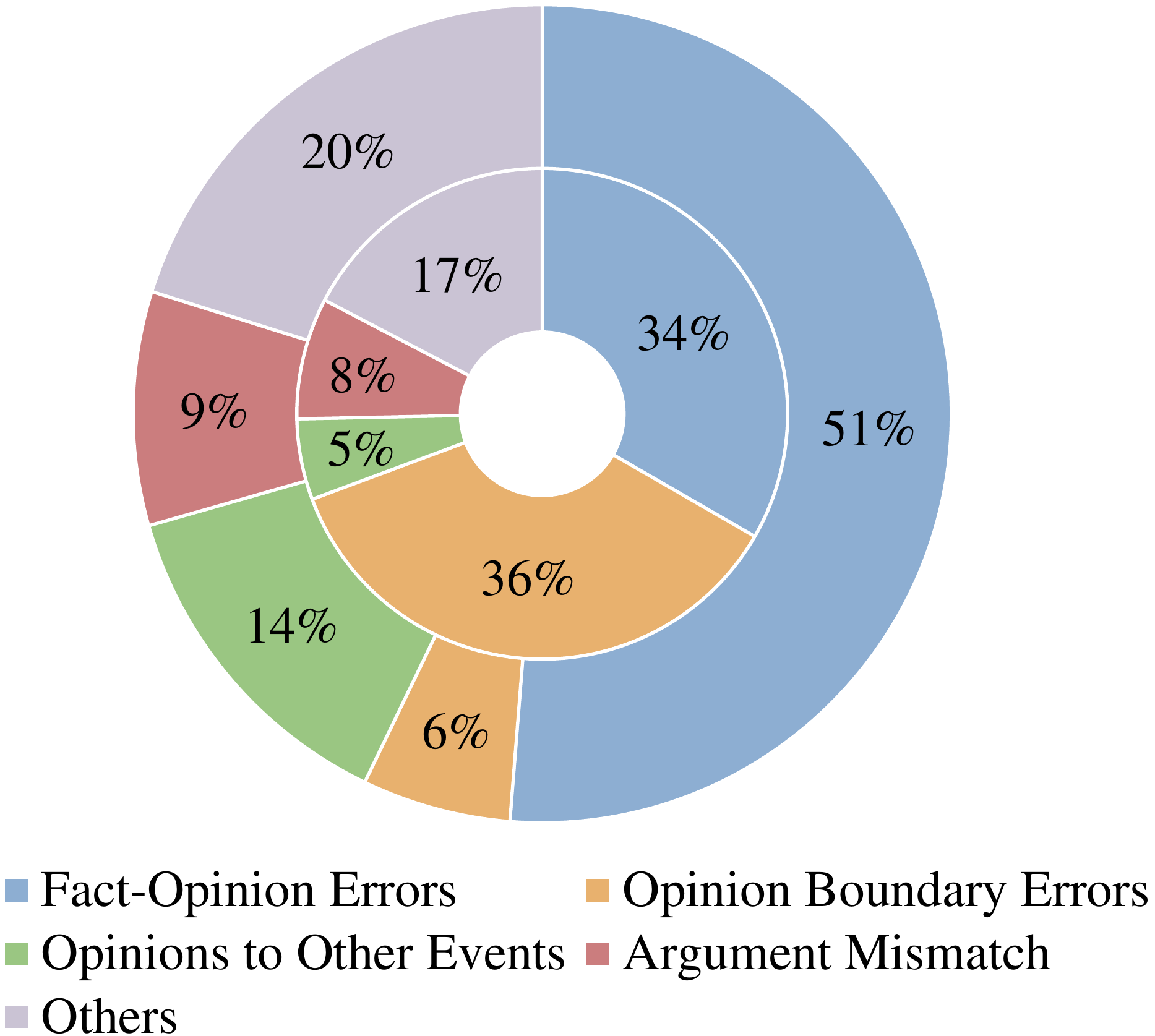} \\
    \caption{Proportions of error causes on ECO of \emph{Seq-MRC}. Inner circle refers to performance on ECO-ZH and outer circle corresponds to ECO-EN dataset.}
    \label{error_causes}
\end{figure}

\subsection{Error Analysis \& Discussion}
To better understand the challenge and bottlenecks of event-centric opinion mining, we further randomly sampled 50 annotated documents from ECO-ZH and ECO-EN respectively. 
We then categorized the errors made by Seq-MRC model to figure out the critical issues to resolve. From Figure \ref{error_causes}, we can see that:

1) The confusions between facts and opinions are one of the most critical EOE errors for both English and Chinese portions. 51\% errors in English and 34\% errors in Chinese portion stem from fact-opinion confusion. This corresponds to the nature of the task because event-centric opinions are frequently mixed up with many facts and non-opinion information. And a sentence can contain both opinion and fact at the same time, which makes it very difficult to identify.

2) Opinion boundary errors are more significant in Chinese portion than English portion. Compared with 6\% boundary errors in English portion, the percentage in Chinese portion is a much higher 36\%. We believe this is because the average length of opinions in Chinese is longer than that in English, which is shown in Table~\ref{dataset}. As a result, more opinion boundary errors are introduced in EOE. Furthermore, by looking into the error cases, we find that such errors mainly occur in cases where two continuous opinions refer to different arguments, which is very challenging.

3) OTE errors are more severe in English portion than Chinese portion. We find that such errors happened more frequently on the opinions with implicit targets.
Furthermore, there are notable 14\% errors in English portion that comes from identifying opinions not corresponding to the given event. This usually happens when models are confused by strong opinion marker words like \emph{say} and \emph{believe}, and similar arguments such as World War I and World War II. We believe that this is because event descriptors in the English portion are much less than Chinese portion. As a result, models overfit on some spurious features and can not sufficiently capture the correct event-oriented information. To alleviate this problem, we will enlarge the English portion of ECO Bank in the future.

\section{Related Work}
Previous opinion mining (OM) researches focus on entity-centric opinions~\cite{liu2007opinion}, which mainly categorizes the holder's sentiments towards entities and their attributes at document-level~\cite{turney2002thumbs,moraes2013document,sharma2014opinion,tang2015document,paredes2017sentiment}, sentence-level~\cite{hatzivassiloglou2000effects,riloff2003learning,hu2004mining,riloff2006feature,sayeed2012grammatical,alessia2015approaches} and aspect-level~\cite{jin2009novel, li2010structure, qiu2011opinion,liu2012sentiment,mitchell2013open,liu2015fine,wang2017coupled, zhao2020spanmlt,peng2020knowing,cai2021aspect,mao2021joint}. 

There are also some researches working on event-related opinions~\cite{karamibekr2012sentiment,zhou2013sentiment,deng2015mpqa,deng2015joint,qian2016multi,gate}. Generally speaking, these studies commonly regard event as a special type of entity, neglecting the unique characteristics of event-centric opinions. However, events are very different from entities, and therefore event-centric opinions have different connotations and targets which have not been exploited yet.

For the evaluation resource of OM, most of current studies are based on the Semeval Challenges datasets~\citep{pavlopoulos2014aspect, pontiki2015semeval, pontiki2016semeval} and its extension~\citep{wang2017coupled, fan2019target, peng2020knowing}, which consist of entity-centric customer reviews about target entities from 7 domains. To the best of our knowledge, the constructed ECO Bank is the first publicly available event-centric opinion mining benchmark from news domain, which definitely can benefit future research in this direction.

\section{Conclusions and Future Work}

In this paper, we propose and formulate \textbf{\textit{event-centric opinion mining}}, a new task that aims to mine a broader range of opinions oriented to specific events from documents. An Event-Centric Opinion Bank corpus is constructed and a two-step framework is proposed. Experiments demonstrate the challenges and advantages of mining event-centric opinions. The focus of this paper is the introduction of the new task and datasets. The proposed four baseline systems are relatively simple and leave much room for further improvements. In future work, we will try to build end-to-end models that directly extract opinion triples in an end-to-end fashion and enrich the current opinion structure.

\section{Ethics Consideration}
In consideration of ethical concerns, we provide the following detailed description:
\begin{enumerate}
    \item All of the collected documents and event descriptors come from publicly available sources. The legal advisor of our institute and/or the original dataset constructor confirms that the sources of our data are freely accessible online without copyright constraint to academic use. 
    \item ECO Bank contains 5000 annotated documents with 988 event descriptors. After double-checking, we guarantee that ECO Bank doesn't contain samples that may cause ethic issues. The dataset does not involve any personal sensitive information. All references in the annotated data are double-checked for plausibility and grammaticality by different human annotators. All documents and event descriptors are also manually checked to ensure they are informative and logically coherent. We manually check the content of each piece of data in ECO Bank to ensure that it does not contain any hate speech or attacks on vulnerable people.
    \item We hired 5 annotators who have bachelor degrees. Before formal annotation, annotators were asked to annotate 20 samples randomly extracted from the dataset, and based on average annotation time we set a fair salary (i.e., 35 dollars per hour) for them. During their training annotation process, they were paid as well.
\end{enumerate}

\section*{Acknowledgement}
We thank all reviewers for their valuable comments. Moreover, this work was supported by the National Key Research and Development Program of China (No. 2020AAA0106400), and the National Natural Science Foundation of China under Grants no. U1936207, 62122077 and 62106251.

\bibliography{anthology,custom}
\bibliographystyle{acl_natbib}

\end{document}